%% file: main.tex
\documentclass[10pt,twocolumn,letterpaper]{article}

\usepackage[pagenumbers]{cvpr} 

\input{preamble}
\definecolor{cvprblue}{rgb}{0.21,0.49,0.74}

\usepackage{pifont}
\usepackage[table, dvipsnames]{xcolor}
\usepackage{multirow}
\usepackage{booktabs}
\usepackage{enumitem}
\usepackage{amsmath}
\usepackage{amsfonts}
\usepackage{amssymb}
\usepackage{colortbl}

\usepackage{makecell}

\usepackage[ruled,vlined]{algorithm2e}
\SetKwInput{KwInput}{Input}
\SetKwInput{KwOutput}{Output}

\newcommand{\detections}{\mathcal{D}}
\newcommand{\tracks}{\mathcal{T}}
\newcommand{\costmatrix}{\mathbf{C}}

\SetKwProg{Proc}{Procedure}{}{}
\SetCommentSty{textit}

\usepackage{float}

\definecolor{tblue}{HTML}{D3CDE4}
\definecolor{alogrblue}{HTML}{5188d1}
\definecolor{alogrgrey}{HTML}{727880}
\definecolor{alogryellow}{HTML}{a07018}
\definecolor{alogrgreen}{HTML}{095f42}
\definecolor{rowgray}{gray}{0.95}
\definecolor{goodgreen}{rgb}{0.0, 0.5, 0.0}
\definecolor{badred}{rgb}{0.7, 0.0, 0.0}
\usepackage[pagebackref,breaklinks,colorlinks,allcolors=cvprblue]{hyperref}

\def\paperID{12766} 
\def\confName{CVPR}
\def\confYear{2026}

\title{DM$^3$T: Harmonizing Modalities via Diffusion for Multi-Object Tracking}

\author{Weiran Li\textsuperscript{1}\hspace{0.2cm}
	Yeqiang Liu\textsuperscript{1}\hspace{0.2cm}
	Yijie Wei\textsuperscript{1}\hspace{0.2cm}
	Mina Han\textsuperscript{1}\hspace{0.2cm}
	Qiannan Guo\textsuperscript{2}\hspace{0.2cm}
	Zhenbo Li\textsuperscript{1}\thanks{Corresponding author.}
	\vspace{1.5mm}\\
	\textsuperscript{\rm 1}China Agricultural University\hspace{0.5cm}
	\textsuperscript{\rm 2}Beijing Normal University
	\vspace{1mm}\\
	{\tt\small vranlee86@gmail.com, lizb@cau.edu.cn}\\
	{\tt\small yeqiangliu@cau.edu.cn, yjwei@cau.edu.cn, hmnsx666@163.com, guoqiannan1203@163.com}
}

\begin{document}
\maketitle
\input{sec/0_abstract}    
\input{sec/1_intro}
\input{sec/2_related}
\input{sec/3_method}
\input{sec/4_exps}
\input{sec/5_conclusion}
\input{sec/X_suppl}

{
	\small
	\bibliographystyle{ieeenat_fullname}
	\bibliography{main}
}

\end{document}

%% file: sec/0_abstract.tex
\begin{abstract}
Multi-object tracking (MOT) is a fundamental task in computer vision with critical applications in autonomous driving and robotics. Multimodal MOT that integrates visible light and thermal infrared information is particularly essential for robust autonomous driving systems. However, effectively fusing these heterogeneous modalities is challenging. Simple strategies like concatenation or addition often fail to bridge the significant non-linear distribution gap between their feature representations, which can lead to modality conflicts and degrade tracking accuracy. Drawing inspiration from the connection between \textbf{M}ulti\textbf{M}odal \textbf{M}O\textbf{T} and the iterative refinement in \textbf{D}iffusion models, this paper proposes \textbf{DM$^3$T}, a novel framework that reformulates multimodal fusion as an iterative feature alignment process to generate accurate and temporally coherent object trajectories. Our approach performs \textit{iterative cross-modal harmonization} through a proposed \textit{Cross-Modal Diffusion Fusion} (C-MDF) module. In this process, features from both modalities provide mutual guidance, iteratively projecting them onto a shared, consistent feature manifold. This enables the learning of complementary information and achieves deeper fusion compared to conventional methods. Additionally, we introduce a plug-and-play \textit{Diffusion Refiner} (DR) to enhance and refine the unified feature representation. To further improve tracking robustness, we design a \textit{Hierarchical Tracker} that adaptively handles confidence estimation. DM$^3$T unifies object detection, state estimation, and data association into a comprehensive online tracking framework without complex post-processing. Extensive experiments on the VT-MOT benchmark demonstrate that our method achieves 41.7 HOTA, representing a 1.54\% relative improvement over existing state-of-the-art methods. The code and models are available at \textit{\href{https://vranlee.github.io/DM-3-T/}{https://vranlee.github.io/DM-3-T/}}.
\end{abstract}

%% file: sec/1_intro.tex
\section{Introduction}
\label{sec:intro}

\begin{figure}[t]
    \centering
    \includegraphics[width=1\linewidth]{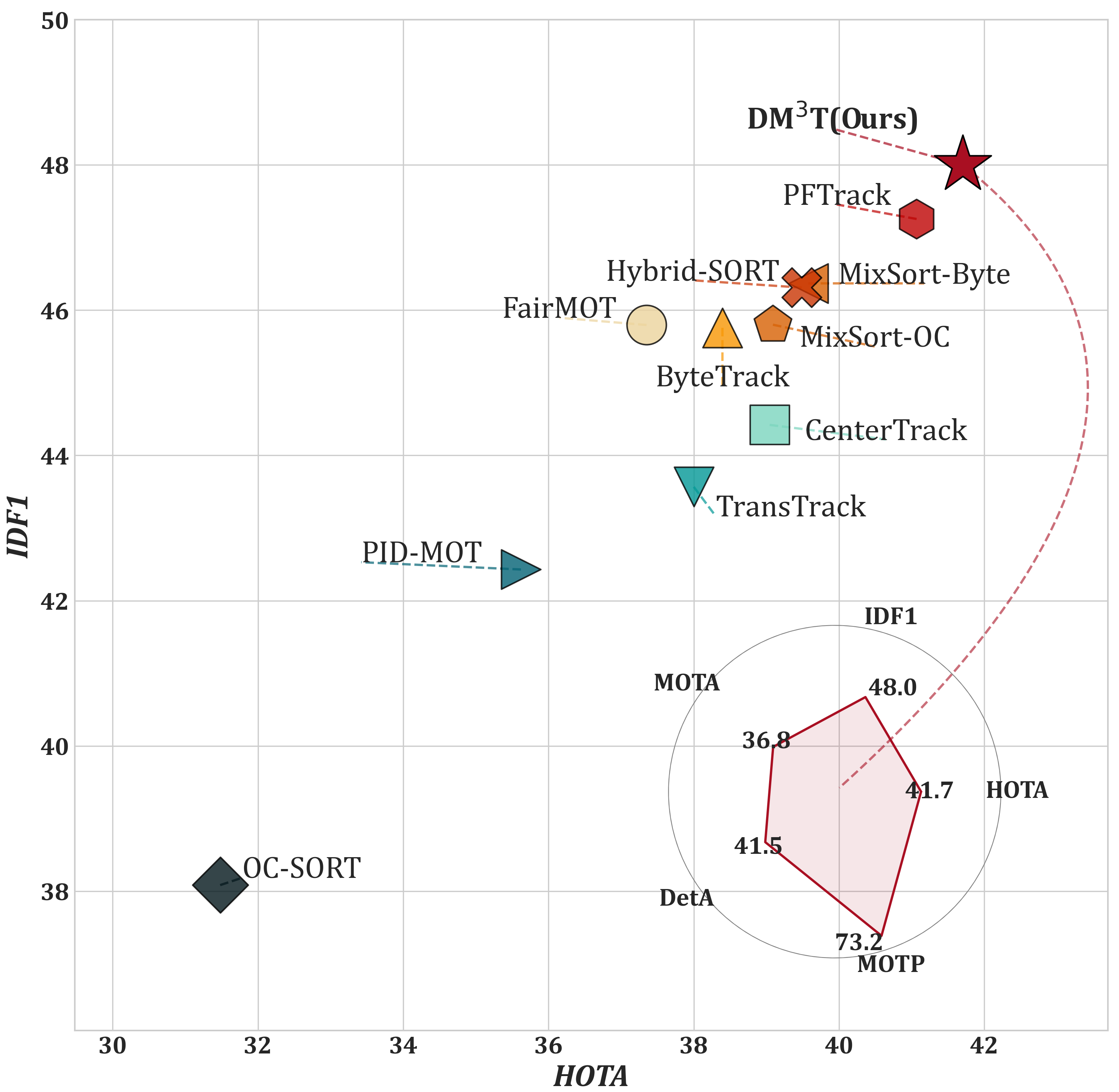}
    \caption{Performance comparison on the VT-MOT test set. Our DM$^3$T achieves state-of-the-art performance with HOTA of 41.7 and IDF1 of 48.0. The subfigure presents results across additional evaluation metrics. Detailed quantitative results are provided in Table~\ref{tab:tracking_performance}.}
    \label{fig:sota}
\end{figure}

Multi-object tracking (MOT) aims to simultaneously localize multiple targets and maintain their consistent identities across video sequences~\cite{hassan2024multi, zhao2022tracking, fischer2023qdtrack}. This fundamental computer vision task is essential for autonomous driving, surveillance systems, and human-computer interaction~\cite{yi2024ucmctrack,liu2023fasttrack,feng2023towards}. Despite significant advances in RGB-based MOT methods, their performance deteriorates substantially under adverse conditions such as low illumination, weather disturbances, and complex lighting scenarios~\cite{liu2023collaborative, liu2025sparsetrack}. These limitations are inherent to visible-light imaging, which fundamentally compromises the reliability of tracking systems in challenging real-world environments.

Thermal infrared imaging offers a compelling solution by capturing long-wave infrared radiation emitted by objects, enabling robust detection of heat-emitting targets even in complete darkness~\cite{zhang2019multi, gao2025tvtracker}. The complementary nature of RGB and thermal modalities has motivated extensive research in RGB-T multimodal fusion, which combines the rich textural and semantic information of RGB imagery with the robust perceptual capabilities of thermal sensing~\cite{gao2025tvtracker, hui2023bridging}. This dual-modality approach promises environment-agnostic tracking performance superior to either modality alone.

However, effectively fusing heterogeneous modalities poses significant challenges~\cite{zhang2024exploring,xiao2022attribute}. Conventional fusion strategies, such as feature concatenation, element-wise operations, and attention mechanisms, are constrained by critical limitations. Their primary failure is an inability to adequately capture the complex, non-linear distribution gap between the two modalities. Simply combining features that exist in different, unaligned manifolds can lead to modality conflicts and sub-optimal representations, limiting the capacity to adaptively emphasize informative cues.

Recent breakthroughs in denoising diffusion probabilistic models (DDPMs)~\cite{ho2020denoising} have achieved impressive generative results by progressively reconstructing data~\cite{croitoru2023diffusion, dhariwal2021diffusion}. This iterative refinement process offers a powerful parallel for multimodal fusion, conceptualizing it as a progressive alignment of heterogeneous data modalities onto a unified, consistent representation, rather than as simply cleaning noisy inputs.

Motivated by this observation, we introduce DM$^3$T, a novel RGB-thermal multi-object tracking framework that reformulates multimodal fusion as an iterative feature alignment process. We design a \textit{Cross-Modal Diffusion Fusion} module that leverages iterative cross-modal harmonization to guide information exchange between RGB and thermal features, enabling complementary enhancement while resolving cross-modal conflicts. We further propose a plug-and-play \textit{Diffusion Refiner} that progressively improves the fused features through diffusion-based refinement. Our complete framework integrates these components into a \textit{Hierarchical Tracker} that achieves robust object association, especially in challenging low-confidence scenarios, as shown in Fig.~\ref{fig:sota}. Our main contributions are as follows:

\begin{itemize}
\item We propose DM$^3$T, a novel diffusion-based RGB-thermal tracking framework that reformulates multimodal fusion as an iterative feature alignment process, providing a new paradigm for cross-modal integration.
\item We design a \textit{Cross-Modal Diffusion Fusion} module that enables iterative cross-modal guidance to harmonize features and enhance complementary information between RGB and thermal modalities.
\item We introduce a plug-and-play \textit{Diffusion Refiner} that progressively improves fused features through diffusion-based refinement, boosting the discriminative power for downstream tracking tasks.
\item We develop a \textit{Hierarchical Tracker} that stabilizes object associations in challenging scenarios by effectively handling low-confidence detections through robust multimodal representations.
\end{itemize}

%% file: sec/2_related.tex
\section{Related Work}
\label{sec:related}

\subsection{Multi-Object Tracking}

Most modern multi-object tracking (MOT) methods follow either the Tracking-By-Detection (TBD)~\cite{wojke2017simple, seidenschwarz2023simple} or Joint-Detection-Embedding (JDE)~\cite{wang2020towards, zhang2021fairmot} paradigm. TBD methods decompose tracking into separate detection and association steps, first obtaining object observations through detectors~\cite{cao2023observation,luo2024diffusiontrack}, then performing tracking via motion models and data association algorithms like the Hungarian algorithm~\cite{zhang2022bytetrack}. However, this separation often leads to suboptimal performance as detection errors propagate to the tracking stage, and the independent design objectives create inconsistencies between components.

In contrast, JDE-based methods~\cite{wang2020towards} and Transformer architectures~\cite{dosovitskiy2020image, carion2020end} use unified networks that jointly optimize detection and embedding extraction for trajectory prediction~\cite{sun2020transtrack}. While these integrated approaches offer better computational efficiency, they often struggle to learn discriminative appearance features for individual objects, limiting their performance in challenging tracking scenarios.

\subsection{Diffusion in Multimodal Vision}

Diffusion models have emerged as a powerful generative framework, inspiring applications that extend far beyond traditional generation tasks~\cite{dhariwal2021diffusion}. In multimodal perception, these models demonstrate remarkable capability in aligning and integrating diverse inputs, such as RGB images and textual descriptions~\cite{yang2024mma}. Recent research conceptualizes diffusion as a conditional process wherein one modality guides the generation or refinement of another~\cite{shuai2024survey}.

At the feature level, diffusion models function as effective cross-modal message passing mechanisms~\cite{zhu2025visible}. These approaches leverage cross-modal attention and mutual refinement to iteratively enhance shared representations. Although several diffusion-based tracking methods have been proposed~\cite{luo2024diffusiontrack, lv2024diffmot}, they predominantly concentrate on detection heads or bounding box generation, thereby constraining their applicability to broader multimodal fusion challenges. Our approach addresses this limitation by reformulating feature-level fusion as an iterative alignment process, inspired by the refinement principles of diffusion models. Through a hierarchical design, we achieve both deep cross-modal interactions and robust representation learning, ultimately delivering improved tracking performance.

\begin{figure*}[ht]
    \centering
    \includegraphics[width=1\linewidth]{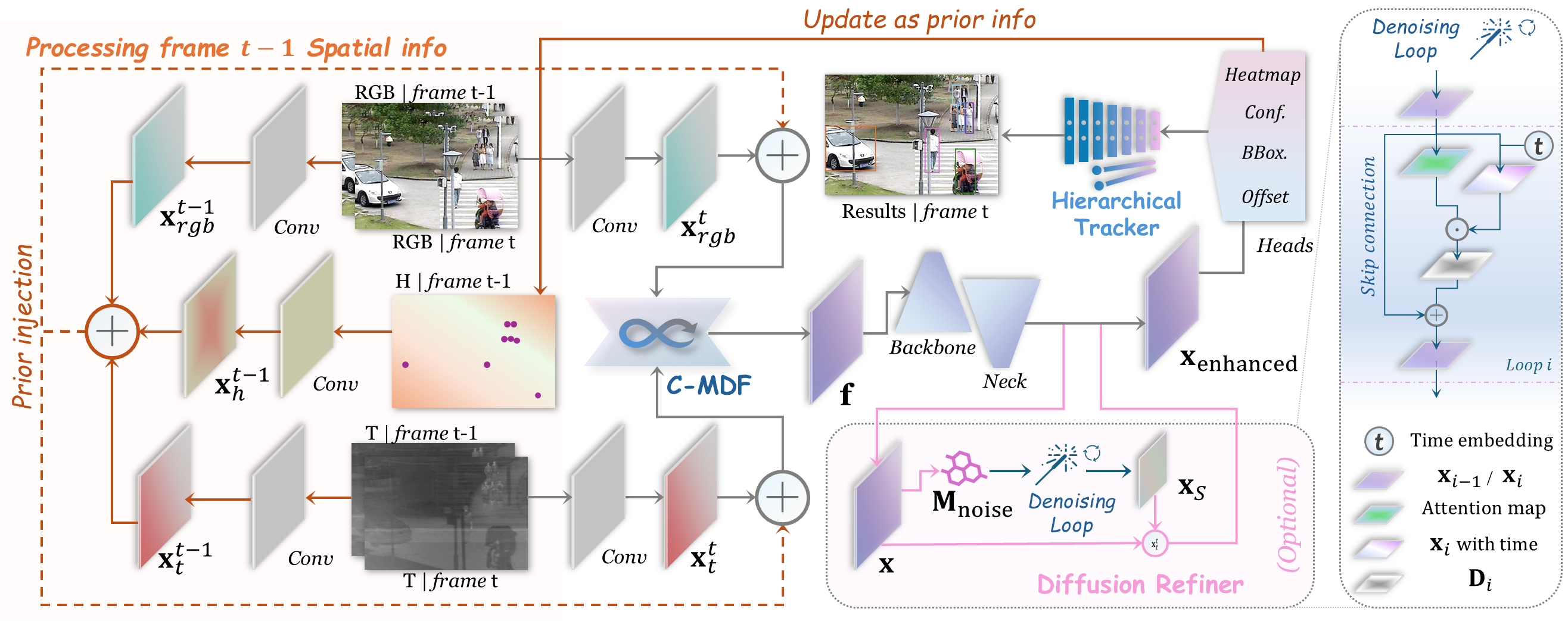}
    \caption{Overall architecture of DM$^3$T. The framework consists of a \textit{Cross-Modal Diffusion Fusion} module that performs iterative cross-modal harmonization between RGB and thermal features, a \textit{Diffusion Refiner} for further feature enhancement, and a \textit{Hierarchical Tracker} that manages object associations through confidence-guided multi-stage association and adaptive motion prediction.}
    \label{fig:framework}
\end{figure*}

%% file: sec/3_method.tex
\section{Methodology}
\label{sec:method}

\subsection{Framework}

Our fundamental insight is rooted in the observation that the iterative refinement mechanisms inherent in diffusion models can be effectively adapted to progressively harmonize multimodal feature representations through bidirectional cross-modal guidance. This paradigm allows each modality to leverage its distinctive strengths while simultaneously compensating for modality-specific limitations, thereby producing more robust and discriminative feature representations for tracking applications.

Building upon this principle, we propose DM$^3$T, a comprehensive online tracking framework that seamlessly integrates complementary information from visible and thermal imagery while preserving object identity consistency through an advanced temporal association strategy. The proposed architecture consists of two primary components: a diffusion-based cross-modal feature extraction and fusion network, and a multi-stage online tracking module. Fig.~\ref{fig:framework} illustrates the complete algorithmic workflow of our framework.

To strengthen temporal coherence, we incorporate historical information ($\mathbf{x}_{\text{h}}$) by integrating image features from the preceding frame and object position priors encoded as heatmap ($\text{H}$) representations into the current frame's feature extraction pipeline. The tracking system takes paired visible-thermal (RGB-T) image sequences as input. After initial feature extraction from each modality, our \textit{Cross-Modal Diffusion Fusion} module performs deep feature integration to generate semantically enriched unified representations. Subsequently, a multi-task detection head simultaneously predicts object positions, dimensions, and motion displacements. Finally, the online association module correlates these detection outputs with existing trajectory hypotheses, ensuring robust and temporally consistent tracking performance.

\subsection{Cross-Modal Diffusion Fusion}

\begin{figure*}
    \centering
    \includegraphics[width=1\linewidth]{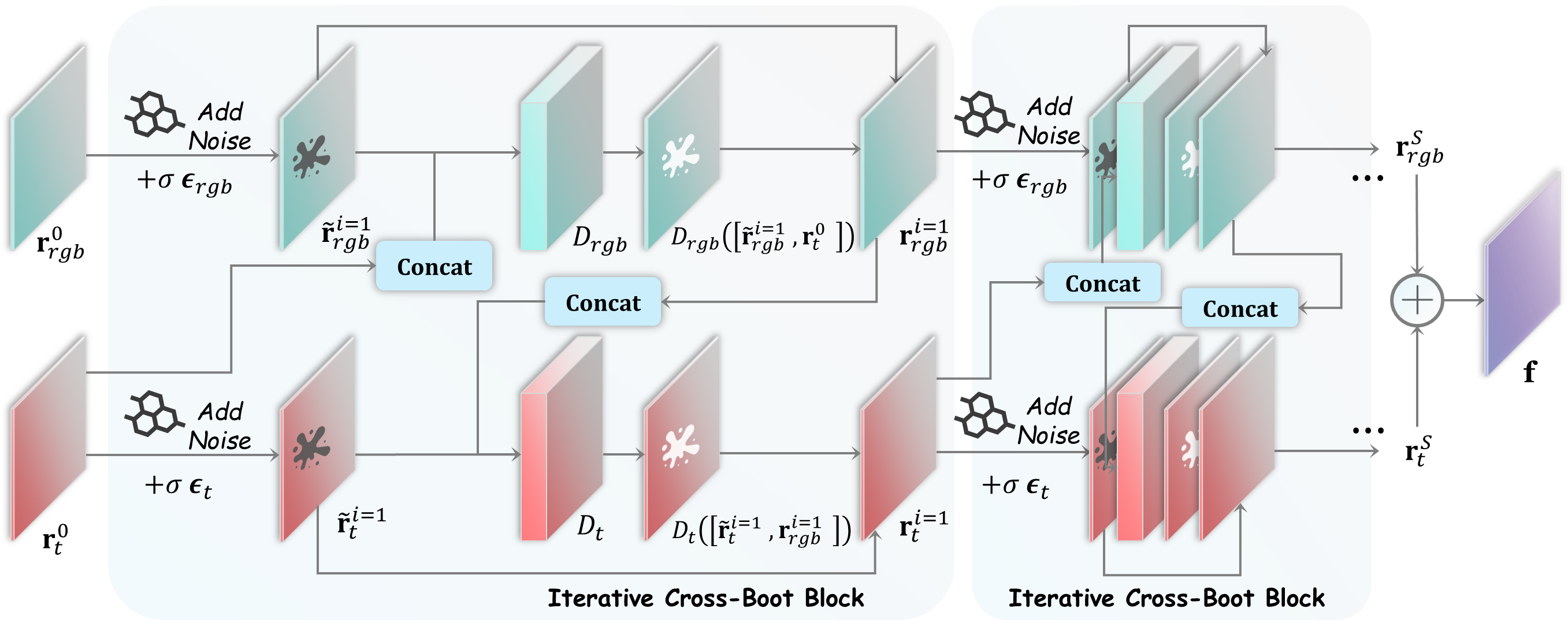}
    \caption{The pipeline of \textit{Cross-Modal Diffusion Fusion} (C-MDF). The module consists of iterative cross-modal harmonization blocks that perform controlled perturbation and refinement to progressively align RGB and thermal features. Each refinement network processes concatenated features from both modalities, enabling deep cross-modal interaction.}
    \label{fig:C-MDF}
\end{figure*}

The core of our approach lies in the \textit{Cross-Modal Diffusion Fusion} (C-MDF) module, which implements an iterative cross-modal harmonization process between RGB and thermal features. Unlike traditional fusion methods that perform single-step integration, our approach enables progressive refinement through multiple stages of mutual interaction. The pipeline of C-MDF is illustrated in Fig.~\ref{fig:C-MDF}. 

\subsubsection{Iterative Cross-Modal Harmonization}

We introduce controlled random perturbations to force the network to learn a more robust shared feature manifold. During iterative cross-guidance, one modality must seek information from another modality to reconstruct the perturbated information, thus achieving deeper feature alignment rather than simply feature stacking.

The C-MDF module is constructed by stacking iterative harmonization blocks. Both modalities employ parallel refinement networks. Formally, given the initial RGB and thermal features $\mathbf{x}_{\text{rgb}}$ and $\mathbf{x}_{\text{t}}$, the module performs the following operations:

\begin{equation}
\mathbf{r}_{rgb}^0 = \mathbf{x}_{rgb}, \mathbf{r}_{t}^0 = \mathbf{x}_{t}
\end{equation}

To facilitate deep cross-modal interaction, the refinement process iteratively applies controlled perturbation and guided harmonization over $S$ steps. At each iteration $i$ ($1 \leq i \leq S$), we first perturb the current modality-specific features:

\begin{equation}
\mathbf{\tilde{r}}_{rgb}^i = \mathbf{r}_{rgb}^{i-1} + \sigma \boldsymbol{\epsilon}_{rgb}
\end{equation}
\begin{equation}
\mathbf{\tilde{r}}_{t}^i = \mathbf{r}_{t}^{i-1} + \sigma \boldsymbol{\epsilon}_{t}
\end{equation}

\noindent where $\sigma$ is a perturbation scaling hyperparameter, and $\boldsymbol{\epsilon}_{rgb}, \boldsymbol{\epsilon}_{t} \sim \mathcal{N}(0, \mathbf{I})$ are random tensors.

Next, the refinement networks for each modality, $D_{rgb}$ and $D_{t}$, take as input the concatenation of the perturbed features from one modality and the previous-step features from the other modality. The refined features are updated as follows:

\begin{equation}
    \mathbf{r}_{rgb}^i = \mathbf{\tilde{r}}_{rgb}^i + D_{rgb}\left([\mathbf{\tilde{r}}_{rgb}^i, \mathbf{r}_{t}^{i-1}]\right)
\end{equation}
\begin{equation}
    \mathbf{r}_{t}^i = \mathbf{\tilde{r}}_{t}^i + D_{t}\left([\mathbf{\tilde{r}}_{t}^i, \mathbf{r}_{rgb}^{\mathbf{i-1}}]\right)
\end{equation}

\noindent where $[\cdot, \cdot]$ denotes channel-wise concatenation. This design enables each modality to iteratively leverage complementary cues from the other, progressively resolving modality-specific conflicts and enhancing discriminative information. We set the number of iteration steps $S=3$.

\subsubsection{Refinement Network Architecture}

The refinement networks $D_{rgb}$ and $D_t$ enable effective cross-modal information exchange within our iterative refinement framework. These networks are specifically designed to process concatenated cross-modal features and generate residual corrections that progressively refine features while enhancing complementary information between modalities.

Each modality-specific refinement network $D_m$ ($m\in\{\text{rgb},\text{t}\}$) is implemented as a lightweight residual convolutional block. Specifically, the network takes the concatenated cross-modal features $[\mathbf{\tilde{r}}_m^i, \mathbf{r}_{n}^{i-1}]$ as input (where $n$ denotes the other modality) and processes them through a sequence of operations: a $3 \times 3$ convolution, batch normalization, ReLU activation, a second batch normalization, and a final $3 \times 3$ convolution. The output is a learned residual, which is added back to the perturbed features to facilitate stable and efficient training.

The concatenation $[\mathbf{\tilde{r}}_m^i, \mathbf{r}_{n}^{i-1}]$ denotes channel-wise fusion of features from both modalities at each iteration, enabling effective cross-modal information exchange. The residual design, combined with batch normalization, enhances gradient flow and stabilizes optimization, while the compact architecture ensures computational efficiency during iterative refinement.

\par After $S$ iterations, the final fused feature representation is obtained by summing the outputs from both modalities:

\begin{equation}
    \mathbf{f} = \mathbf{r}_{rgb}^S + \mathbf{r}_{t}^S
\end{equation}

\noindent where $\mathbf{f}$ serves as the unified cross-modal feature for downstream tracking tasks.

\subsection{Diffusion Refiner}

While C-MDF effectively integrates complementary information across modalities, the resulting features may still contain residual inconsistencies and can benefit from further refinement to enhance discriminative details critical for tracking. To address this challenge, we introduce a \textit{Diffusion Refiner} (DR), a plug-and-play module that applies diffusion-inspired principles to progressively refine and enhance the unified feature representation. This module leverages an iterative, diffusion-inspired process to explore the feature space, yielding more robust representations. It refines the fused features $\mathbf{f}$ using a time-conditioned refinement network, an attention mechanism, and residual connections to preserve critical tracking information while enhancing discriminative details.

\subsubsection{Iterative Refinement Process}

A key aspect of our DR is the incorporation of time embeddings, which provide the network with awareness of the current refinement stage. Given an input feature map $\mathbf{x} \in \mathbb{R}^{C \times H \times W}$, we generate a time embedding $\mathbf{t}_{\text{emb}}$ for step $t$ using:

\begin{equation}
    \mathbf{t}_{\text{emb}} = \text{MLP}(t/T)
\end{equation}

\noindent where $T$ is the total diffusion steps and MLP is a two-layer perceptron with SiLU activation. This embedding is spatially expanded and concatenated with the feature map to form time-conditioned features.

The refinement follows an iterative approach over $S$ steps. Initially, we apply controlled perturbation:

\begin{equation}
    \mathbf{x}_{\text{perturbed}} = \mathbf{x} + \sigma \cdot \mathbf{M}_{\text{noise}}(\mathbf{x}) \cdot \boldsymbol{\epsilon}
\end{equation}

\noindent where $\mathbf{M}_{\text{noise}}(\mathbf{x})$ predicts adaptive perturbation scales, $\boldsymbol{\epsilon} \sim \mathcal{N}(0, \mathbf{I})$ is Gaussian noise, and $\sigma$ controls overall perturbation level. For each step $i$ from 1 to $S$, we perform:

\begin{equation}
    \mathbf{x}_i = \mathbf{D}_i(\mathbf{x}_{i-1}, t_i) + \mathbf{S}_i(\mathbf{x}_{i-1})
\end{equation}

\noindent where $\mathbf{D}_i$ is the $i$-th refinement block and $\mathbf{S}_i$ is a skip connection. Specifically, the block $\mathbf{D}_i$ leverages an internal attention mechanism to re-weight the features, allowing the model to dynamically focus on more informative spatial regions during the refinement. Concurrently, the skip connection $\mathbf{S}_i$ ensures that essential information from the previous step is directly passed forward, preventing feature degradation during the iterative process. The term $t_i = 1 - i/S$ represents the normalized diffusion timestep. The refined features are integrated with the original input via a controlled residual connection:

\begin{equation}
    \mathbf{x}_{\text{enhanced}} = \mathbf{x} + \alpha_R \cdot (\mathbf{x}_S - \mathbf{x})
\end{equation}

\noindent where $\alpha_R$ balances preservation of original features with introduction of refined details, enabling our model to maintain critical tracking information while enhancing discriminative features.

\subsection{Hierarchical Tracker}

While our diffusion-based feature fusion significantly improves detection quality, robust tracking requires a sophisticated association strategy to handle the inherent uncertainties of multimodal perception. To this end, we introduce a hierarchical tracking framework that systematically manages detections of varying confidence levels. The process follows the standard \textit{Predict-Associate-Update} paradigm and is detailed in Algorithm~\ref{alg:hierarchical_tracker}.

\subsubsection{Confidence-Guided Multi-Stage Association}

A key challenge in tracking is that even well-fused features can produce detections with a wide range of confidence scores. Our tracker addresses this through a multi-stage, confidence-guided association strategy that prioritizes high-certainty matches.

Our approach processes detections in descending order of their confidence scores. We establish a sequence of decreasing confidence thresholds, $\tau_1 > \tau_2 > \dots > \tau_k$. Specifically, we define $k=6$ thresholds. At each stage $i$, we consider only the subset of detections $\mathcal{D}_i = \{d \in \mathcal{D} \mid \text{score}(d) > \tau_i\}$ that have not yet been matched. This subset is then associated with the set of currently unmatched tracks $\mathcal{T}_{\text{unmatched}}$. By processing high-confidence detections first, we ensure that the most reliable information is used to update existing tracks, preventing potential mismatches. Detections and tracks matched in a given stage are removed from subsequent, lower-confidence stages. This hierarchical process ensures that high-quality associations are secured first, while still allowing the tracker to leverage lower-confidence detections to maintain tracks that might otherwise be lost.

\subsubsection{Adaptive Motion and State Prediction}

Our framework employs an adaptive motion model to handle temporary occlusions and maintain trajectory continuity.

\noindent\textbf{1. Predict:} Before association, we first predict the new position for each existing track. For a track $t$ with position $\vec{p}_{t-1}$ and velocity $\vec{v}_{t-1}$ from the previous frame, we predict its current position $\hat{\vec{p}}_t$ using a simple constant velocity model:
\begin{equation}
\hat{\vec{p}}_t = \vec{p}_{t-1} + \vec{v}_{t-1}
\label{eq:prediction}
\end{equation}
This predicted position $\hat{\vec{p}}_t$ is used during the association stage to compute the cost matrix against new detections.

\noindent\textbf{2. Update:} If a track is successfully matched with a new detection $d$ (with center $\vec{p}_d$), we update the track's state. The track's new position $\vec{p}_t$ is set to the detection's position $\vec{p}_d$. We then update its velocity $\vec{v}_t$ using an exponential moving average (EMA) to smooth the motion:
\begin{equation}
\vec{v}_{t} = (1-\alpha)\vec{v}_{t-1} + \alpha(\vec{p}_{t} - \vec{p}_{t-1})
\label{eq:velocity_update}
\end{equation}
where $\vec{p}_{t-1}$ is the track's position in the previous frame, and $\alpha$ is the smoothing factor, which we set to 0.7 to prioritize recent measurements. The track's age is reset to $a_t = 0$.

\noindent\textbf{3. Handle Unmatched:} For tracks that remain unmatched after all association stages, we increment their age $a_t \leftarrow a_t + 1$. We then apply a velocity decay mechanism to temper the influence of prediction over time, preventing unbounded error accumulation. The velocity is adjusted as:
\begin{equation}
\vec{v}_{t} \leftarrow \vec{v}_{t-1} \cdot \max(0.5, 1.0 - \beta \cdot a_t)
\label{eq:velocity_decay}
\end{equation}
\noindent where $a_t$ is the number of frames since the track's last successful match, and the decay rate $\beta$ is set to 0.1. This ensures that the predictive drift is gracefully reduced. Tracks with $a_t > a_{\text{max}}$ are terminated.

\subsubsection{Multi-Cue Association Metric}

To accurately associate detections with tracks, we employ a comprehensive cost metric. The association cost $C(d, t)$ between a detection $d$ and a track $t$ is a weighted sum of spatial distance and scale (size) dissimilarity. The primary component is the squared Euclidean distance between the detection center $\vec{p}_d$ and the track's predicted center $\hat{\vec{p}}_t$ (from Eq.~\ref{eq:prediction}):

\begin{equation}
D_{\text{spatial}}(d, t) = \|\vec{p}_d - \hat{\vec{p}}_t\|^2
\label{eq:spatial_dist}
\end{equation}

To this, we add a penalty for significant differences in the scale of the bounding boxes:

\begin{equation}
D_{\text{size}}(d, t) = \frac{|\text{size}(d) - \text{size}(t)|}{\max(\text{size}(d), \text{size}(t))}
\label{eq:size_dist}
\end{equation}

The final cost matrix $\mathbf{C}$ is formulated as:

\begin{equation}
C(d,t) = D_{\text{spatial}}(d,t) + \lambda_s D_{\text{size}}(d,t)
\label{eq:cost_metric}
\end{equation}

\noindent where $\lambda_s=100$ is a weighting factor. To ensure correctness, we enforce two hard constraints. First, associations are only permitted between detections and tracks of the same semantic class. Second, we apply a gating mechanism, invalidating any potential match where the spatial distance $D_{\text{spatial}}(d,t)$ exceeds a dynamic threshold proportional to the track's size. These constraints are implemented by adding a very large penalty to the cost matrix for invalid pairings. The final association problem is solved using the Hungarian algorithm.

\begin{algorithm}[h!]
    \caption{Hierarchical Tracker}
    \label{alg:hierarchical_tracker}
    
    \KwInput{Detections $\detections$, Tracks $\tracks$ (state $\mathbf{p}_{t-1}, \mathbf{v}_{t-1}, \text{age}$)}
    \KwOutput{Updated tracks $\tracks'$}
    \BlankLine
    
    Sort $\detections$ by confidence (descending)\;
    Set thresholds $\mathcal{K} = \{\tau_1 > \dots > \tau_k\}$\;
    $\tracks_{U} \leftarrow \tracks$; $\detections_{U} \leftarrow \detections$; $\tracks_{M} \leftarrow \emptyset$\;
    \BlankLine
    
    \textcolor{alogrgrey}{/* 1. Predict locations */}\\
    \ForEach{track $t \in \tracks$}{
        $\mathbf{p}'_t \leftarrow \mathbf{p}_{t-1} + \mathbf{v}_{t-1}$ (Eq.~\ref{eq:prediction})\;
        Update $t$'s bounding box\;
    }
    \BlankLine
    
    \textcolor{alogrgrey}{/* 2. Multi-stage association */}\\
    \ForEach{threshold $\tau_i \in \mathcal{K}$}{
        $\detections_i \leftarrow \{d \in \detections_{U} \mid \text{score}(d) > \tau_i\}$\;
        \If{$\detections_i = \emptyset$}{\KwSty{continue}\;}
        \If{$\tracks_{U} = \emptyset$}{\KwSty{break}\;}

        $\costmatrix \leftarrow$ ComputeCost($\detections_i, \tracks_{U}$) (Eq.~\ref{eq:cost_metric})\;
        $\text{matches} \leftarrow \text{SolveAssignment}(\costmatrix)$\;
        
        \ForEach{matched pair $(d, t) \in \text{matches}$}{
            Update $t$: $\mathbf{p}_t \leftarrow \mathbf{p}_d$, Update $\mathbf{v}_t$ (Eq.~\ref{eq:velocity_update}), $\text{age} \leftarrow 0$\;
            Add $t$ to $\tracks_{M}$\;
            Remove $d$ from $\detections_{U}$ and $t$ from $\tracks_{U}$\;
        }
    }
    \BlankLine
    
    \textcolor{alogrgrey}{/* 3. Handle new tracks */}\\
    \ForEach{$d \in \detections_{U}$ \KwSty{where} $\text{score}(d) > \tau_{\text{new}}$}{
        Initialize new track $t$ from $d$ ($\text{age} \leftarrow 0, \mathbf{v}_t \leftarrow [0,0]$)\;
        Add $t$ to $\tracks_{M}$\;
    }
    \BlankLine
    
    \textcolor{alogrgrey}{/* 4. Handle unmatched tracks */}\\
    \ForEach{$t \in \tracks_{U}$}{
        \If{$\text{age}(t) < a_{\text{max}}$}{
            $\text{age}(t) \leftarrow \text{age}(t) + 1$\;
            Decay $\mathbf{v}_t$ (Eq.~\ref{eq:velocity_decay})\;
            Update $t$'s position\;
            Add $t$ to $\tracks_{M}$\;
        }
    }
    \BlankLine
    
    $\tracks' \leftarrow \tracks_{M}$\;
    \KwRet{$\tracks'$}\;
    
\end{algorithm}

%% file: sec/4_exps.tex
\section{Experiments}
\label{sec:exps}

\subsection{Experimental Setting}

We validate our approach on the large-scale VT-MOT benchmark~\cite{zhu2025visible} using an NVIDIA A100 GPU. We employ the standard HOTA~\cite{luiten2021hota}, IDF1, and CLEAR metrics~\cite{dendorfer2021motchallenge}, with HOTA as our primary metric. Our model is trained for 10 epochs using the Adam optimizer with a learning rate of $1.25 \times 10^{-4}$ and a batch size of 32, following standard protocols for fair comparison.

\subsection{Ablation Studies}

\subsubsection{Diffusion Fusion and Refinement}

\setlength{\tabcolsep}{0.8mm}
\begin{table*}[htbp]
    \small
    \centering
    \caption{Ablation study of the proposed \textit{Cross-Modal Diffusion Fusion} (C-MDF) and \textit{Diffusion Refiner} (DR) modules. The baseline (\textit{Base}) uses simple concatenation. We report the performance change relative to this baseline.}
    \label{tab:ablation_diffusion}
    \begin{tabular}{l|cc|ccc|ccccccc}
    \toprule
    \textbf{Method} & \textbf{DR} & \textbf{C-MDF} & \textbf{HOTA$\uparrow$} & \textbf{IDF1$\uparrow$} & \textbf{Params$\downarrow$} & \textbf{MOTA$\uparrow$} & \textbf{AssA$\uparrow$} & \textbf{DetA$\uparrow$} & \textbf{IDs$\downarrow$} & \textbf{IDFP$\downarrow$} & \textbf{IDFN$\downarrow$} & \textbf{Frag$\downarrow$} \\
    \midrule
    
    Base & & & 38.633 & 44.259 & \textbf{16.747M} & \textbf{38.962} & 41.338 & 38.280 & \textbf{9799} & 239719 & 526185 & 10389 \\
    
    \rowcolor{tblue!50}
    & & \multirow{2}{*}{} & 38.871 & 43.948 & 16.808M & 36.889 & 41.968 & 37.871 & 10335 & \textbf{238705} & 529203 & 9939 \\
    
    \rowcolor{tblue!50}
    \multirow{-2}{*}{Refining Only} & \multirow{-2}{*}{\ding{51}}& & \textcolor{goodgreen}{+0.238 {\tiny(+0.61\%)}} & \textcolor{badred}{-0.311 {\tiny(-0.70\%)}} & \textcolor{badred}{+0.061M {\tiny(+0.36\%)}} & \textcolor{badred}{-2.073} & \textcolor{goodgreen}{+0.630} & \textcolor{badred}{-0.409} & \textcolor{badred}{+536} & \textcolor{goodgreen}{-1014} & \textcolor{badred}{+3018} & \textcolor{goodgreen}{-450} \\
    
    \rowcolor{tblue}
    & & & \textbf{40.635} & \textbf{46.426} & 17.307M & 38.115 & \textbf{41.984} & \textbf{40.948} & 9895 & 317330 & \textbf{483330} & \textbf{9331} \\
    
    \rowcolor{tblue}
    \multirow{-2}{*}{Cross \& Refining} & \multirow{-2}{*}{\ding{51}} & \multirow{-2}{*}{\ding{51}} & \textcolor{goodgreen}{+2.002 {\tiny(+5.18\%)}} & \textcolor{goodgreen}{+2.167 {\tiny(+4.90\%)}} & \textcolor{badred}{+0.560M {\tiny(+3.34\%)}} & \textcolor{badred}{-0.847} & \textcolor{goodgreen}{+0.646} & \textcolor{goodgreen}{+2.668} & \textcolor{badred}{+96} & \textcolor{badred}{+77611} & \textcolor{goodgreen}{-42855} & \textcolor{goodgreen}{-1058} \\
    
    \bottomrule
    \end{tabular}
\end{table*}

\setlength{\tabcolsep}{3.8mm} 
\begin{table*}[htbp]
    \small
\centering
\caption{Ablation of the association strategy. We compare our \textit{Hierarchical} tracker against a \textit{Base} greedy association and the \textit{BYTE} association method. All methods use the same input features from our C-MDF and DR modules.}
\begin{tabular}{l|cc|ccccccc}
\toprule
\textbf{Method} & \textbf{HOTA$\uparrow$} &  \textbf{IDF1$\uparrow$} &\textbf{MOTA$\uparrow$} & \textbf{AssA$\uparrow$} & \textbf{DetA$\uparrow$} & \textbf{IDs$\downarrow$} & \textbf{IDFP$\downarrow$} & \textbf{IDFN$\downarrow$} & \textbf{Frag$\downarrow$} \\
\midrule
Base & 40.635 & 46.426 & 38.115 & 41.984 & 40.948 & 9895  & 317330 & 483330 & 9331 \\
BYTE & 40.108 & 46.791 & \textbf{38.266} & \textbf{43.786} & 38.026 & \textbf{3225}  & \textbf{227280} & 507272 & \textbf{5147} \\
\rowcolor{tblue}
\textbf{Hierarchical}  & \textbf{41.703} & \textbf{48.000} & 36.759 & 43.653 & \textbf{41.461} & 7195  & 341664 & \textbf{460173} & 9749 \\
\bottomrule
\end{tabular}
\label{tab:ablation_hierarchical_tracker}
\end{table*}

We validate the contributions of our core fusion modules, C-MDF and DR, in Table~\ref{tab:ablation_diffusion}. The baseline (\textit{Base}) replaces both modules with simple feature concatenation.

Interestingly, applying only the DR (\textit{Refining Only}) yields minimal HOTA gain and slightly degrades IDF1 (-0.311). This result suggests that the DR module, when applied to raw, un-harmonized features, may inadvertently amplify modality-specific noise. We hypothesize it struggles to refine features that are not yet aligned on a shared manifold.

In stark contrast, the full \textit{Cross \& Refining} model, which first aligns features with C-MDF, yields substantial gains, boosting HOTA by +2.00 and IDF1 by +2.17. This strongly demonstrates that the components work synergistically. The C-MDF module is essential for iterative feature alignment, forcing the network to learn a unified representation by resolving cross-modal conflicts. The DR module can then successfully operate on this harmonized manifold to enhance discriminative details. The significant improvement in detection (DetA +2.67) and the large reduction in false negatives (IDFN -42k) validate our pipeline's effectiveness in generating robust representations for tracking.

\subsubsection{Association}

We then evaluate our \textit{Hierarchical Tracker} against a \textit{Base} associator and the strong \textit{BYTE}~\cite{zhang2022bytetrack} associator (Table~\ref{tab:ablation_hierarchical_tracker}). All methods use identical input features from our C-MDF+DR modules.

The \textit{BYTE} associator excels at reducing IDs and IDFP, consistent with its design. However, this comes at the cost of a significantly higher False Negative rate (IDFN 507k) and lower detection quality (DetA 38.02), leading to a lower overall HOTA score. Our Hierarchical approach, which systematically processes detections by confidence, achieves the best HOTA (41.70) and IDF1 (48.00). It strikes a more effective balance, demonstrating a superior ability to leverage the robust features from our fusion pipeline. It successfully utilizes low-confidence but valid detections (a benefit of C-MDF) that \textit{BYTE} discards, which is critical in challenging multimodal scenarios.

\subsection{Real-time Performance}
\label{sec:runtime_analysis}

We analyze the efficiency-accuracy trade-off of our fusion modules in Table~\ref{tab:runtime_efficiency}. Due to their iterative nature, our modules introduce a modest parameter increase (+0.56M) and a notable GFLOPs overhead (+41.58 GFLOPs). This costs only a minor 2.77 FPS reduction in speed. Crucially, this demonstrates a highly favorable accuracy-to-speed trade-off: we achieve a significant +2.00 HOTA gain while maintaining a real-time speed of 15.31 FPS. This result quantitatively substantiates our framework's suitability for real-time online tracking.

\setlength{\tabcolsep}{2.4mm} 
\begin{table}[htbp]
    \centering
    \caption{Real-time performance trade-off. We compare the baseline (w/o C\&R) against our Cross \& Refining model (w/ C\&R). Speed is the mean $\pm$ std. dev. over 3 runs on the VT-MOT test set.}
    \label{tab:runtime_efficiency}
    \small
    \begin{tabular}{l|c|>{\columncolor{tblue}}c>{\columncolor{tblue}}c}
        \toprule
        \textbf{Metric} & \textbf{w/o C\&R} & \textbf{w/ C\&R} & \textbf{$\Delta$} \\ 
        \midrule
        HOTA$\uparrow$ & 38.633 & \textbf{40.635} & \textcolor{goodgreen}{+2.00} \\
        Params$\downarrow$ & 16.747 M & 17.307 M & \textcolor{badred}{+0.56 M} \\
        GFLOPs$\downarrow$ & 66.279 G & 107.860 G & \textcolor{badred}{+41.58 G} \\
        \midrule
        Speed (FPS)$\uparrow$ & \textbf{18.08 $\pm$ 0.07} & 15.31 $\pm$ 0.66 & \textcolor{badred}{-2.77} \\
        \bottomrule
    \end{tabular}
\end{table}

\subsection{Benchmark Results}

\setlength{\tabcolsep}{1.3mm} 
\begin{table}[htbp]
    \small
    \centering
    \caption{Performance comparison of DM$^3$T with state-of-the-art methods on the VT-MOT benchmark. The best results are highlighted in bold.}
    \label{tab:tracking_performance}
    \begin{tabular}{l|cc|ccc}
    	\toprule
    	\textbf{Method} &  \textbf{HOTA$\uparrow$}  & \textbf{IDF1$\uparrow$} &  \textbf{MOTA$\uparrow$}  &  \textbf{DetA$\uparrow$}  &  \textbf{MOTP$\uparrow$}  \\ \midrule
    	FairMOT         &     37.350      &    45.795     &     37.266      &     34.628      &     72.525      \\
    	CenterTrack     &     39.045      &     44.420     &     30.585      &     38.104      &     72.874      \\
    	TransTrack      &     38.000     &    43.567     &     36.156      &     35.711      &     73.823      \\
    	ByteTrack       &     38.393      &    45.757     &     33.151      &     32.122      &     73.483      \\
    	OC-SORT         &     31.479      &    38.086     &     28.948      &     25.244      &      73.150      \\
    	MixSort-OC      &     39.090     &    45.799     &      31.330      &     33.109      &     73.632      \\
    	MixSort-Byte    &     39.575      &    46.367     &     31.593      &     34.806      &     73.049      \\
    	PID-MOT         &     35.621      &     42.430     &     33.333      &     33.245      &     71.794      \\
    	Hybrid-SORT     &     39.485      &    46.310     &     31.074      &     34.619      &     72.840      \\
    	PFTrack         &     41.068      &    47.254     & \textbf{43.088} & \textbf{41.631} & \textbf{73.949} \\ \midrule
	\rowcolor{tblue}\textbf{DM$^3$T(Ours)}    & \textbf{41.703} & \textbf{48.000} &     36.759      &     41.461      &      73.150      \\ \bottomrule
    \end{tabular}
\end{table}
\setlength{\tabcolsep}{6pt}

As shown in Table~\ref{tab:tracking_performance}, our DM$^3$T achieves state-of-the-art performance, attaining the highest HOTA (41.70) and IDF1 (48.00) scores, surpassing all listed methods including PFTrack~\cite{zhu2025visible}, Hybrid-SORT~\cite{yang2024hybrid}, and ByteTrack~\cite{zhang2022bytetrack}.

Notably, our MOTA (36.76) is lower than PFTrack's (43.09). Our analysis (with Table~\ref{tab:ablation_hierarchical_tracker}) indicates this is a direct trade-off. Our \textit{Hierarchical Tracker} achieves the lowest False Negative (IDFN) rate (460k), significantly boosting HOTA and IDF1. However, this comes at the cost of higher False Positives (IDFP) and Identity Switches (IDs), which negatively impact MOTA. This demonstrates a clear trade-off: our method prioritizes comprehensive detection and association (low IDFN) to achieve the best overall tracking quality (HOTA) and identity preservation (IDF1), at the expense of precision metrics that define MOTA.

\subsection{Cross-Platform}

To evaluate robustness, we analyze performance across the three capture platforms in Table~\ref{tab:Platform}. Our method excels in stable \textit{Surveillance} scenes (51.50 HOTA), where our feature alignment pipeline can effectively leverage complementary information. Performance degrades on dynamic platforms like \textit{Handheld} (39.99 HOTA) and \textit{UAV} (40.21 HOTA), which suffer from camera instability, motion blur, and perspective distortion. This indicates that while our C-MDF module enhances robustness against modality gaps (e.g., darkness), severe camera motion remains a significant challenge, suggesting a need for future work in motion-robust feature extraction or explicit camera motion compensation.

\setlength{\tabcolsep}{3.6mm} 
\begin{table}[htbp]
	\small
	\centering
    \caption{Performance breakdown by capture platform (Handheld, Surveillance, UAV) on the VT-MOT dataset. All metrics reported are the mean values across all sequences for each platform.}
	\label{tab:Platform}
	\begin{tabular}{l|ccc}
		\toprule
		\textbf{Metric}& \textbf{Handheld} & \textbf{Surveillance} & \textbf{UAV} \\
		\midrule
		\textbf{HOTA$\uparrow$} & 39.989 & \textbf{51.502} & 40.214 \\
		\textbf{IDF1$\uparrow$} & 45.664 & \textbf{63.613} & 48.191 \\
		\textbf{MOTA$\uparrow$} & 50.546 & \textbf{60.175} & 29.211 \\
		\textbf{AssA$\uparrow$} & 32.868 & \textbf{51.652} & 45.758 \\
		\textbf{DetA$\uparrow$} & 48.455 & \textbf{53.632} & 37.092 \\
		\textbf{IDs$\downarrow$} & 64.588 & \textbf{51.057} & 62.176 \\
		\textbf{IDFP$\downarrow$} & 2737.667 & \textbf{1797.314} & 4092.265 \\
		\textbf{IDFN$\downarrow$} & 3407.941 & \textbf{1739.400} & 6632.029 \\
		\textbf{Frag$\downarrow$} & 86.588 & \textbf{64.857} & 90.088 \\
		\bottomrule
	\end{tabular}
\end{table}

\subsection{Fusion}
\label{sec:qual_fusion}

\begin{figure}[htbp]
    \centering
    \includegraphics[width=\linewidth]{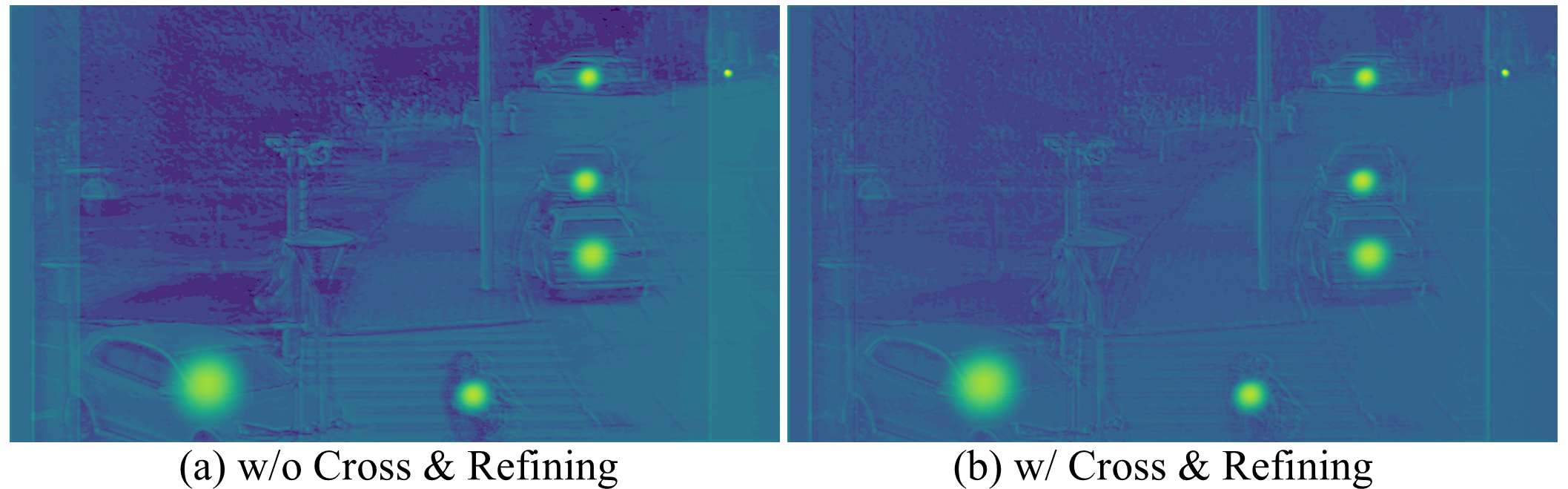}
    \caption{Qualitative comparison of feature maps, with corresponding quantitative analysis in Table~\ref{tab:qual_metrics}. Additional tracking visualizations are provided in the \textit{Supplementary Material}.}
    \label{fig:qual_fusion}
\end{figure}

We validate our fusion pipeline both visually and quantitatively. As shown in Fig.~\ref{fig:qual_fusion}(a), the feature map w/o C\&R processing is visibly corrupted by background clutter and spurious activations, which are artifacts of unresolved modality conflicts. In contrast, the C\&R-processed feature map in Fig.~\ref{fig:qual_fusion}(b) effectively suppresses this noise, producing a clean map with sharp, focused activations. This visually confirms our diffusion-inspired approach successfully harmonizes heterogeneous features and generates a more robust, discriminative representation.

This visual assessment is corroborated by Table~\ref{tab:qual_metrics}. Our model significantly reduces Shannon entropy, indicating less randomness and noise. Conversely, it increases kurtosis, which quantitatively confirms a more peaked distribution with activations concentrated on targets. The lower mean intensity and noise mean further demonstrate the framework's effectiveness in suppressing background clutter. This combined qualitative and quantitative evidence provides strong validation for our core hypothesis.

\setlength{\tabcolsep}{4mm}
\begin{table}[htbp]
    \small
    \centering
    \caption{Quantitative analysis of feature maps shown in Fig.~\ref{fig:qual_fusion}.}
    \label{tab:qual_metrics}
    \begin{tabular}{l|c>{\columncolor{tblue}}c}
        \toprule
        \textbf{Metric} & \textbf{w/o C\&R} & \textbf{w/ C\&R} \\
        \midrule
        Information Entropy (bits)$\downarrow$ & 3.2506 & \textbf{2.5934} \\
        Kurtosis (Peakedness)$\uparrow$ & 117.6522 & \textbf{127.9850} \\
        Noise Mean (1-50)$\downarrow$ & 4.9262 & \textbf{4.2252} \\
        Mean Intensity$\downarrow$ & 5.1638 & \textbf{4.8571} \\
        Std. Deviation$\downarrow$ & 8.2272 & \textbf{7.9963} \\
        \bottomrule
    \end{tabular}
\end{table}

%% file: sec/5_conclusion.tex
\section{Conclusion}
\label{sec:conclusion}

We proposed DM$^3$T, a novel framework that reformulates RGB-T MOT fusion as an iterative feature alignment process. The \textit{Cross-Modal Diffusion Fusion} module harmonizes heterogeneous modalities onto a shared manifold, achieving 41.7 HOTA on the VT-MOT benchmark. However, we identified a critical trade-off: this strategy increases false positives and identity switches, consequently sacrificing the MOTA score. Future work will explore constraining the diffusion refinement process to explicitly model detection uncertainty, aiming to learn a feature representation that better disentangles low-confidence true positives from hard false positives.

%% file: sec/X_suppl.tex
\section*{Appendix.A.~Qualitative Visualization}

\begin{figure*}
	\centering
	\includegraphics[width=1\linewidth]{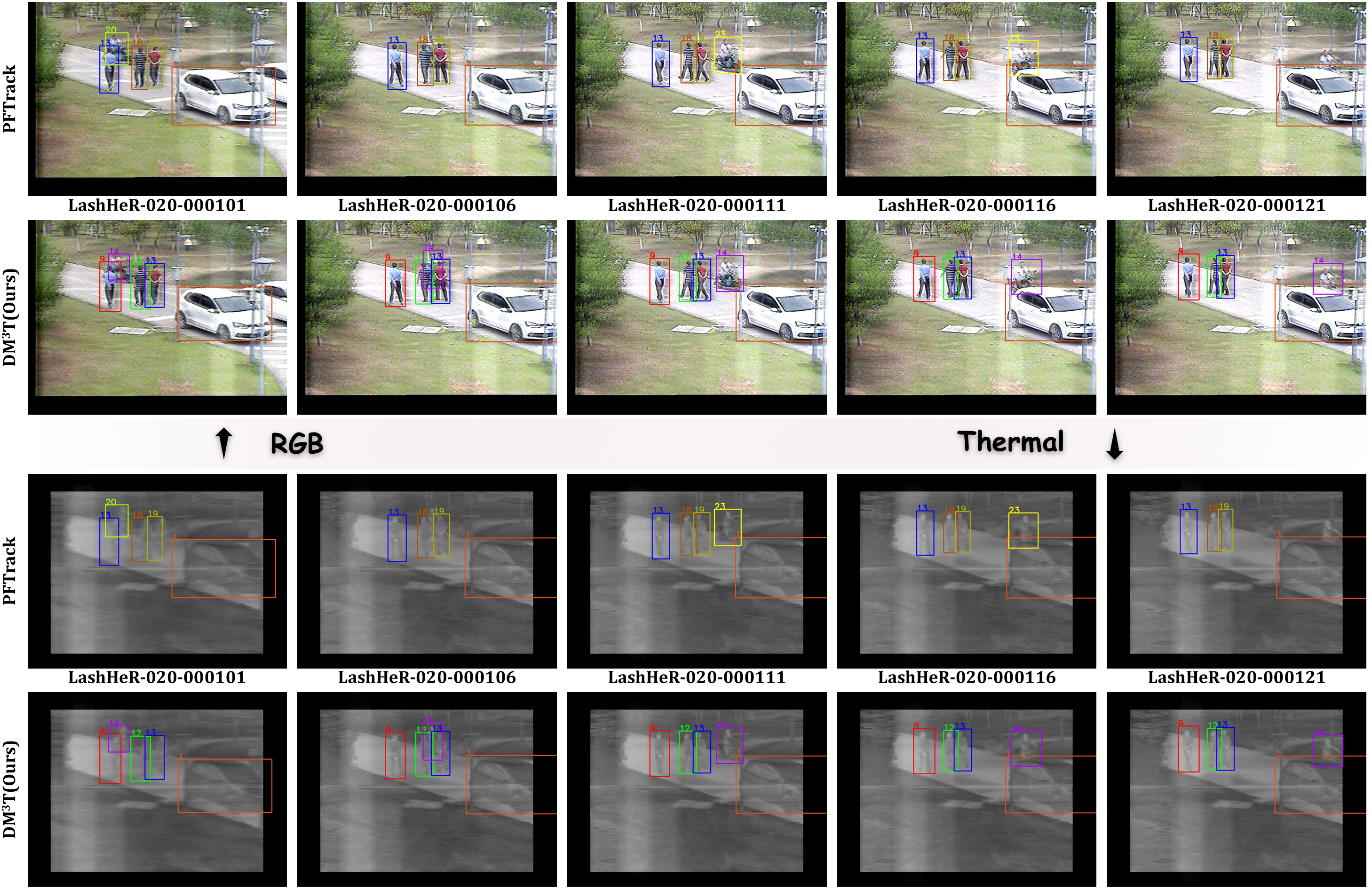}
	\caption{Tracking performance on LashHer-020 sequence. Our method maintains robust multi-object tracking despite significant occlusion. Different object identities are distinguished by bounding box colors and ID numbers, best viewed in color and zoomed. Same as below.}
	\label{fig:LashHer}
\end{figure*}

\begin{figure*}
	\centering
	\includegraphics[width=1\linewidth]{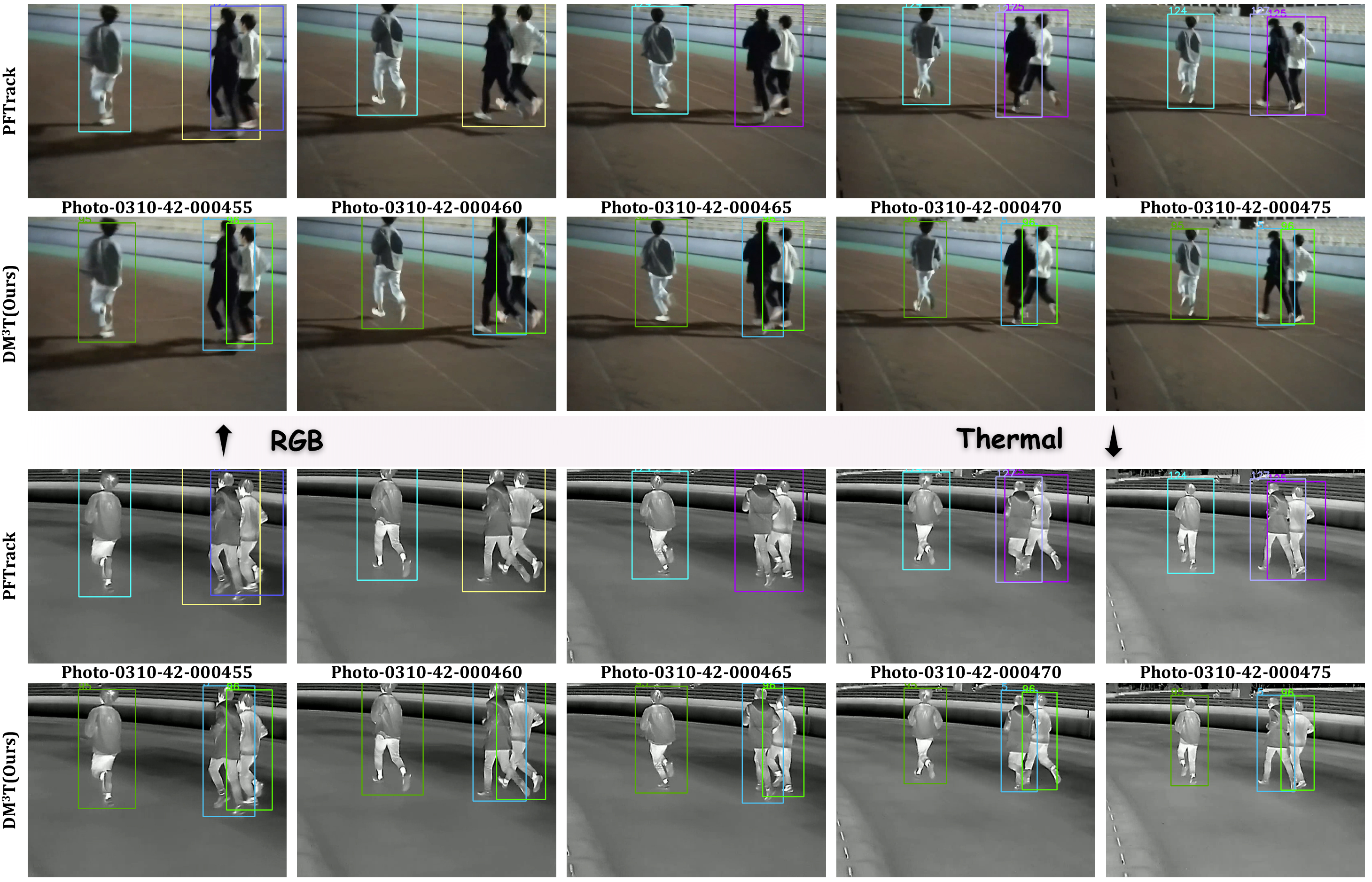}
	\caption{Comparative tracking performance on Photo-0310-42 sequence. Our method (bottom) remains stable despite camera jitter and object adhesion.}
	\label{fig:Photo}
\end{figure*}

We present qualitative comparisons against the strong baseline, PFTrack, in Fig.~\ref{fig:LashHer} and Fig.~\ref{fig:Photo}. In the LashHer-020 sequence (Fig.~\ref{fig:LashHer}), PFTrack fails to handle occlusion effectively, resulting in target loss and incorrect identity reassignment. In contrast, our method maintains consistent identity tracking throughout the entire sequence. This visual evidence corroborates the superior identity preservation capabilities (reflected in our high IDF1 score) enabled by the discriminative, harmonized features generated by our C-MDF module, which demonstrates significant robustness to temporary occlusion.

In the Photo-0310-42 sequence (Fig.~\ref{fig:Photo}), which is characterized by severe camera jitter and object adhesion, our tracker remains stable, whereas PFTrack exhibits frequent identity switches. This further demonstrates that our iterative refinement process successfully resolves cross-modal conflicts and mitigates noise, producing a stable and temporally coherent feature representation that is resilient to error.

\section*{Appendix.B.~Network Architecture Details}

To facilitate reproducibility and ensure a fair comparison with existing methods, we provide comprehensive architectural specifications for our proposed approach. We adopt the widely-used DLA-34 backbone~\cite{yu2018deep} due to its proven effectiveness across MOT benchmarks. This choice allows for an unbiased evaluation of our diffusion-based contributions, independent of potential gains from more complex backbone architectures. Note that our framework maintains backbone agnosticism and can readily accommodate alternative architectures.

\subsection*{B.1.~Backbone: DLA-34}

DLA-34 leverages recursive and hierarchical connections, enabling the retention of both fine-grained spatial details and high-level semantic cues. This design is particularly advantageous for object detection and tracking tasks where multi-scale representations are critical.

Our implementation of DLA-34 comprises six hierarchical stages. The initial levels (Level 0 and Level 1) consist of conventional convolutional layers, while the deeper levels (Levels 2--5) adopt a tree-structured topology implemented using the \textit{Tree} class. These stages utilize the Hierarchical Deep Aggregation (HDA) mechanism to incrementally aggregate features from varying depths, constructing a robust feature hierarchy. The detailed configuration is summarized in Table~\ref{tab:dla34_backbone}.
Each tree module consists of \textit{Basic Block} units following the classic residual design with two $3\times3$ convolutions. This structure ensures stable training and efficient gradient flow, rendering the backbone highly suitable for real-time detection and tracking pipelines.

\subsection*{B.2.~Neck: Iterative Deep Aggregation Upsampling}

To recover spatial resolution and fuse multi-scale features from the backbone, we employ a feature aggregation neck composed of \textit{DLAUp} and Iterative Deep Aggregation Upsampling (\textit{IDAUp}) modules. Inspired by the original DLA design, these modules progressively integrate low-resolution, semantically rich features with high-resolution, spatially detailed maps.

The fusion process originates from the deepest backbone output (Level 5) and iteratively aggregates features up to Level 2. Each \textit{IDAUp} module aligns the channel dimensions of its inputs, upsamples the lower-resolution maps via transposed convolutions, and fuses them through element-wise addition. Detailed architectural specifications are provided in Table~\ref{tab:neck_structure}.

To enhance geometric adaptability, all projection and aggregation operations within the \textit{IDAUp} modules utilize Deformable Convolutions (DCNv2)~\cite{zhu2019deformable}. This capability allows the model to dynamically adjust sampling locations based on object shapes and motion patterns, which is crucial for handling complex multi-object tracking scenarios. The final output of the neck is a unified feature map with a stride of 4 relative to the input image, serving as the input for the detection and tracking heads.

\begin{table*}[htbp]
	\centering
	\small
	\caption{Structure of the DLA-34 backbone. Levels 2--5 are constructed with \textit{Tree} modules employing \textit{Basic Block} as the internal unit.}
	\label{tab:dla34_backbone}
	\begin{tabular}{l|c|c|c|c}
		\toprule
		\textbf{Level} & \textbf{Module} & \textbf{Blocks/Convs} & \textbf{Output Channels} & \textbf{Output Stride} \\
		\midrule
		base\_layer & Conv7x7 & 1 & 16 & 1 \\
		Level 0 & Conv3x3 & 1 & 16 & 1 \\
		Level 1 & Conv3x3 & 1 & 32 & 2 \\
		Level 2 & Tree & 1 & 64 & 4 \\
		Level 3 & Tree & 2 & 128 & 8 \\
		Level 4 & Tree & 2 & 256 & 16 \\
		Level 5 & Tree & 1 & 512 & 32 \\
		\bottomrule
	\end{tabular}%
\end{table*}

\begin{table*}[htbp]
	\centering
	\small
	\caption{Structure of the feature aggregation neck. Each \textit{IDAUp} module merges features from two adjacent levels. All projection and aggregation operations utilize Deformable Convolutions (DCNv2). $L_i$ denotes the feature map from Level $i$. 'd' indicates channel depth.}
	\label{tab:neck_structure}
	\begin{tabular}{l|c|c|c}
		\toprule
		\textbf{Fusion Stage} & \textbf{Inputs} & \textbf{Upsample Operation} & \textbf{Output} \\
		\midrule
		ida\_0 & $L_5$ (512-d), $L_4$ (256-d) & Project $L_5$, Deconv $\times2$, Add & Fused $L_4$ (256-d) \\
		ida\_1 & Fused $L_4$ (256-d), $L_3$ (128-d) & Project, Deconv $\times2$, Add & Fused $L_3$ (128-d) \\
		ida\_2 & Fused $L_3$ (128-d), $L_2$ (64-d) & Project, Deconv $\times2$, Add & Fused $L_2$ (64-d) \\
		\hline
		\textbf{Final Output} & \multicolumn{3}{c}{Fused $L_2$ (64-d, stride 4)} \\
		\bottomrule
	\end{tabular}%
\end{table*}